\pdfoutput=1

\documentclass[11pt]{article}

\usepackage[preprint]{acl}

\usepackage{times}
\usepackage{latexsym}

\usepackage[T1]{fontenc}

\usepackage[utf8]{inputenc}

\usepackage{microtype}

\usepackage{inconsolata}

\usepackage{graphicx}
\usepackage{algorithm}
\usepackage{stackengine}
\usepackage{cite}
\usepackage{color}
\usepackage{xcolor}
\usepackage{amsmath,amsthm,amssymb,amsfonts,bm}
\usepackage{textcomp}
\usepackage{multirow}
\usepackage{url}
\usepackage{makecell}
 \usepackage{booktabs}

\def\BibTeX{{\rm B\kern-.05em{\sc i\kern-.025em b}\kern-.08em
    T\kern-.1667em\lower.7ex\hbox{E}\kern-.125emX}}

\usepackage[noend]{algpseudocode}

\makeatletter
\newcommand*{\rom}[1]{\expandafter\@slowromancap\romannumeral #1@}

\def \v x{\bm x}

\def \v x{\bm X}

\renewcommand{\v}[1]{\ensuremath{\boldsymbol{#1}}}

\title{LAGO: Few-shot Crosslingual Embedding Inversion Attacks via\\ Language Similarity-Aware Graph Optimization   }

\author{Wenrui Yu\textsuperscript{1},\space
  Yiyi Chen\textsuperscript{2},\space
  Johannes Bjerva\textsuperscript{2},\space
  Sokol Kosta\textsuperscript{1},\space
  Qiongxiu Li\textsuperscript{1}\thanks{\ \ Corresponding author.}\\
  \textsuperscript{1}Department of Electronic Systems,\textsuperscript{2}Department of Computer Science\\ 
  Aalborg University, Copenhagen, Denmark\\
\texttt{wenyu@es.aau.dk,\{yiyic,jbjerva\}@cs.aau.dk,\{sok,qili\}@es.aau.dk}
  }

\begin{document}
\maketitle
\begin{abstract}

We propose \textbf{LAGO} - \textit{\textbf{L}anguage Similarity-\textbf{A}ware \textbf{G}raph \textbf{O}ptimization} - a novel approach for few-shot cross-lingual embedding inversion attacks, addressing critical privacy vulnerabilities in multilingual NLP systems. 
Unlike prior work in embedding inversion attacks that treat languages independently, LAGO explicitly models linguistic relationships through a graph-based constrained distributed optimization framework. 
By integrating syntactic and lexical similarity as edge constraints, our method enables collaborative parameter learning across related languages. 
Theoretically, we show this formulation generalizes prior approaches, such as ALGEN, which emerges as a special case when similarity constraints are relaxed. 
Our framework uniquely combines Frobenius-norm regularization with linear inequality or total variation constraints, ensuring robust alignment of cross-lingual embedding spaces even with extremely limited data (as few as 10 samples per language). 
Extensive experiments across multiple languages and embedding models demonstrate that LAGO  substantially improves the transferability of attacks with $10$-$20\%$ increase in Rouge-L score over baselines. This work establishes language similarity as a critical factor in inversion attack transferability, urging renewed focus on language-aware privacy-preserving multilingual embeddings.
\end{abstract}

\section{Introduction}
Text embeddings, which encode semantic and syntactic information into dense vector representations, serve as the backbone of modern natural language processing (NLP) systems.
They are also powering the Large Language Models, whose impact stretches far beyond NLP and is steadily shaping everyday lives and business operations.
However, their widespread deployment in cloud-based services introduces significant privacy risks.
A particularly concerning threat is the embedding inversion attack~\citep{10.1145/3372297.3417270,chen2025algen}, where the adversaries can decode sensitive and private data directly from the embedding vectors. 
The security of the system can be compromised when malicious users abuse the embedding model API, collecting massive datasets to train attack models. Data leakage, whether accidental or deliberate, further exacerbates this vulnerability.
As vector databases and generative-AI services proliferate across the globe, the embedding vectors offered as commodities are mostly multilingual. 
Yet, prior researches in this attack space mostly concentrate on inverting English embeddings~\citep{10.1145/3372297.3417270, li-etal-2023-sentence, morris2023text, huang_transferable_2024}. 
While recent efforts~\citep{chen2024typ,chen2024text, chen2025algen} touch upon multilingual and cross-lingual inversion attacks, they \textit{lack an explicit modeling of language similarities}, resulting in poor generalization across languages.
\begin{figure}[t]
    \centering
\includegraphics[width=0.5\textwidth]{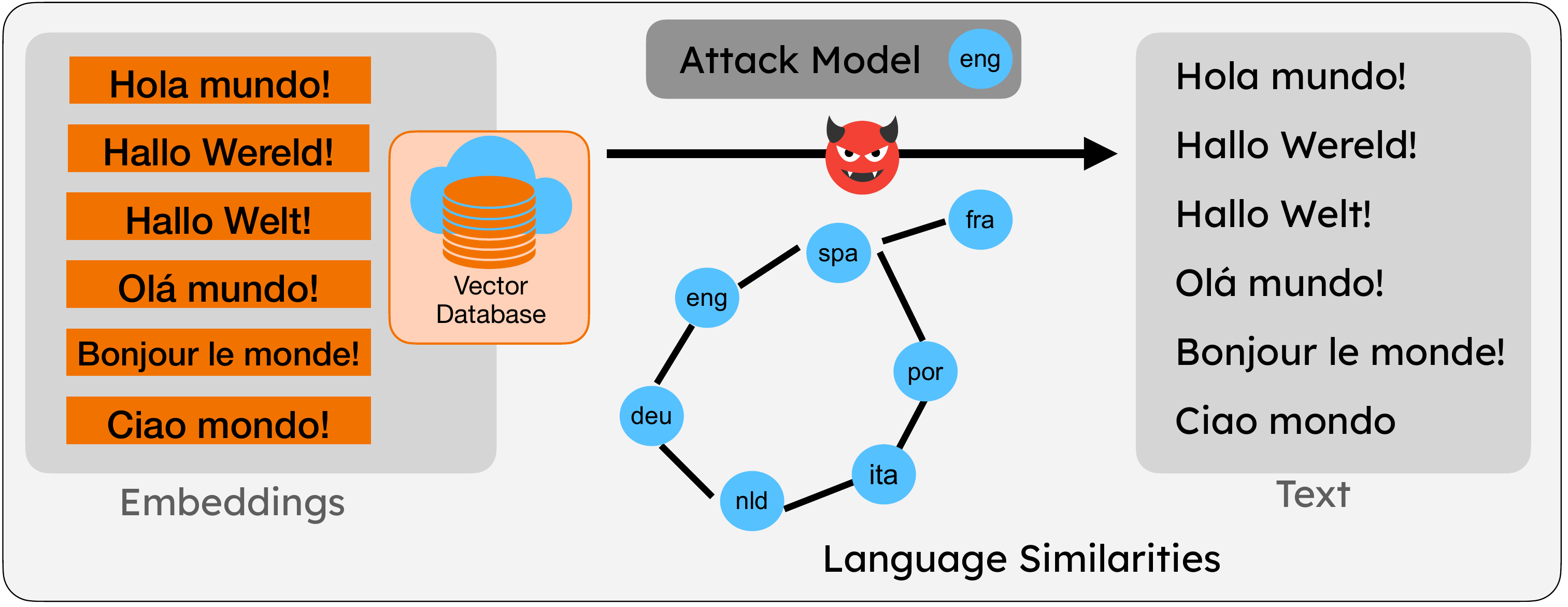}
    \caption{Few-shot Cross-lingual Textual Embedding Inversion Leveraging Language Similarities. Example: Attack model trained on English embeddings is used to attack embeddings in other languages, using language similarities as a prior. }
    \label{fig:figure1}
\end{figure}

In real-world adversarial scenarios, such as specialized domains or low-resource languages, attackers may only have access to a handful of embedding-text pairs.  
Although ALGEN~\citep{chen2025algen} partially addresses the few-shot regime through decoder transfer, it lacks mechanisms to exploit language similarity, which we hypothesize is a key factor in crosslingual generalization failure. Prior studies have shown that language similarities, simulated from typological features and lexical overlapping, correlate with structural variations in inversion outputs~\citep{chen2024typ, chen-etal-2025-large}, providing empirical motivation for incorporating such relationships into attack models.

To address this, we propose \textbf{LAGO} (\textit{\textbf{L}anguage Similarity-\textbf{A}ware \textbf{G}raph \textbf{O}ptimization}) for few-shot crosslingual embedding inversion. LAGO explicitly models linguistic relationships by constructing a topological graph over languages, where nodes represent languages and edges encode similarity. As illustrated in Fig.~\ref{fig:figure1}, this graph is used to guide collaborative optimization of decoder alignment functions across languages, enabling knowledge transfer from typologically related neighbors. We formalize the attack objective as a distributed optimization problem, where each node corresponds to a language and constraints encode similarity-based consistency. We present two algorithmic variants: (1) an inequality-constrained approach based on IEQ-PDMM~\citep{heusdens2024distributed}, and (2) a total variation regularized formulation \citep{peng2021byzantine} that softly penalizes parameter drift across similar languages. Our main contributions include:
\begin{itemize}
    \item We propose LAGO, the first framework for few-shot crosslingual embedding inversion that incorporates language similarity as a structural prior in a graph-constrained optimization problem.
    \item We develop two algorithmic variants, one using inequality constraints and one using total variation penalties, that enable collaborative parameter learning across languages. Prior work, including ALGEN, emerges as a special case within our framework (cf. Section~\ref{ssec:equ}).
    \item Experiments across multiple embedding models and diverse languages show that language similarity strongly correlates with attack transferability, improving performance by 10 - 20\% over prior methods.
\end{itemize}
By exposing overlooked vulnerabilities in multilingual embedding systems and demonstrating effective inversion under realistic low-resource conditions, our work underscores the urgent need for stronger privacy protections in cross-lingual NLP deployments. While differential privacy offers some protection against our attack, it also significantly degrades downstream utility~\citep{chen2025algen}, highlighting the need for more targeted and efficient defense mechanisms.

\section{Related Work}
\subsection{Embedding Inversion Attacks}
Early work on embedding inversion framed the task as classification over fixed vocabularies. For example, \citet{10.1145/3372297.3417270} aim to recover input tokens directly from embeddings, achieving up to 70\% reconstruction. Subsequent advances recast the task as generation: \citet{li-etal-2023-sentence} introduce a decoder-based approach to produce fluent text, while \citet{morris2023text} further improve accuracy through iterative refinement. Several work have since then extended inversion attacks to multilingual scenarios~\citep{chen2024text, chen2024typ}. 
Moreover, \citet{huang_transferable_2024} trains a surrogate model to conduct transfer attack on victim embeddings under black-box access.

These methods, however, typically rely on massive training samples (8k to 5 million victim embeddings) and are primarily evaluated in monolingual or well-resourced settings.  
In practice, attackers often face few-shot scenarios.  For example, reconstructing text in low-resource languages or specialized domains with only a handful of available samples.  ALGEN~\citep{chen2025algen} introduces a linear alignment technique, allowing a decoder trained in one domain to be reused in another. While effective in few-shot transfer, ALGEN does not explicitly model language similarity or structural relationships between languages (cf. Section~\ref{algen}). Our framework improves ALGEN by directly incorporating linguistic knowledge to achieve stronger few-shot cross-lingual inversion.

\subsection{Cross-lingual Transferability}
Crosslingual transferability is an central research topic in multilingual NLP. 
Prior researches leverage crosslingual transferability to improve downstream task performances in target languages, mainly through fine-tuning LLMs on related source languages~\citep{choenni2023languages}, or using zero-shot transfer~\citep{adelani-etal-2022-masakhaner,  de-vries-etal-2022-make, blaschke2025analyzing} or few-shot trasnfer with pre-trained MLLMs~\citep{Lauscher2020FromZT}.
Language similarity based on linguistic data, such as typological features~\citep{littell2017uriel} and lexical databases~\citep{wichmann2022asjp}, have been used extensively in facilitating crosslingual transfer~\citep{philippy2023towards}.
In this work, we leverage language similarity generated from both syntactic features and lexical overlap to provide alternative perspectives on constructing graphs, to assist crosslingual inversion attacks. 

\subsection{Distributed Optimization}
Distributed optimization decomposes a global objective into smaller local problems that are solved collaboratively across networked nodes. Owing to its scalability and efficiency, it has become a foundational tool in large-scale machine learning and signal processing. Applications span domains such as federated learning~\citep{mcmahan2017communication}, sensor networks~\citep{rabbat2004distributed}, and privacy-preserving systems~\citep{li2020privacy, yu2024provable}. Classical distributed optimization algorithms include the Alternating Direction Method of Multipliers (ADMM, \citep{boyd_distributed_2010}) and Primal-Dual Method of Multipliers (PDMM, \citep{zhang2017distributed}) and their variants \citep{wang2014bregman,ouyang2015accelerated,heusdens2024conic,heusdens2024distributed}.  To the best of our knowledge, their application to inversion attacks remains unexplored. In this work, we present, for the first time, a novel migration of distributed optimization techniques to inversion attacks.

\section{Preliminaries}

\subsection{Embedding Inversion attack}

Let $x\in \mathbb{V}^s$ denote a sequence of text tokens, and the text encoder $\phi =enc(\cdot): \mathbb{V}^s\rightarrow \mathbb{R}^n $ be an embedding function that maps text $x$ to a fixed-length vector $\phi(x)\in\mathbb{R}^n$. $s$ is the
sequence length and $n$ the embedding dimension respectively. An embedding inversion attack is formally defined as the process of learning an approximate inverse function $g=dec(\cdot)$ such that:
\vspace{-4pt}  
\[
g(\phi(x)) \approx x\,.
\vspace{-4pt} 
\]

\subsection{ALGEN}\label{algen}
ALGEN enables cross-domain and cross-lingual sentence-level inversion through a framework combining embedding alignment and sequence generation. The framework consists of three parts: 
\paragraph{1) Training a local attack model \(dec_{A}(\cdot)\)} by fine-tuning a pre-trained decoder to function as an embedding-to-text generator.
\paragraph{2) Embedding Alignment} To bridge the discrepancy between the victim \( \v{e}_V \in \mathbb{R}^m\) and the attack \( \v{e}_A \in \mathbb{R}^n\) embedding spaces, a linear mapping matrix \( \v W  \in \mathbb{R}^{m \times n}\) is learned:
\vspace{-5pt}  
    \[
    \v{\hat{e}}_A = \v{e}_V \v W \quad.
    \vspace{-5pt}  
    \]

The optimal alignment matrix \( \v W \) is obtained by solving the following least-squares minimization:  
\vspace{-5pt}  
\[
\min_{\v W} \left\| \v{E}_{A} - \v{E}_{V} \v W \right\|_F^2,
\vspace{-5pt}  
\] 
where \( \|\cdot\|_F \) denotes the Frobenius norm, \( \v{E}_V = [\v e_V^{1\top},\cdots,\v e_V^{b\top}]^\top\in\mathbb{R}^{b\times m} \) is the victim model’s embedding matrix, and \( \v{E}_A = [\v e_A^{1\top},\cdots,\v e_A^{b\top}]^\top \in\mathbb{R}^{b\times n} \) is the attacker’s embedding matrix, and \(b\) is the number of training samples. \(m \) and \(n\) are the embedding dimensions of the victim and attack models, respectively. This optimization problem admits a closed-form solution via the normal equation (see the derivation in Appendix~\ref{normal_equation}):  
\vspace{-6pt}  
\[
\v W = (\v{E}_V^\top \v{E}_V)^{-1} \v{E}_V^\top \v{E}_A,
\vspace{-2pt}  
\]  
which minimizes the reconstruction error between the aligned victim embeddings \( \v{E}_V \v W \) and the attacker’s reference embeddings \( \v{E}_A \).  

\paragraph{3) Text Reconstruction} The aligned embeddings \( \mathbf{\hat{e}}_A \) are decoded into text via \( dec_A \), i.e.,
\vspace{-4pt}  
    \[
    \hat{x} = dec_A(\v{\hat{e}}_A) = dec_A(\v{e}_V \v W).
    \vspace{-4pt}  
    \]

ALGEN achieves inversion without retraining the victim model, requiring only fine-tuning of \( dec_A \) and estimation of \( \v W \).

\subsection{Fundamentals of Distributed Optimization}
Distributed optimization addresses global optimization problems through a unified objective function while incorporating constraints derived from inter-node relationships within the network. Formally, this approach can be expressed as
\begin{align*}
\vspace{-3pt}  
\min_{\{w_i:i \in \mathcal{V}\}} ~~~~&\displaystyle \sum_{i \in \mathcal{V}} f_i(w_i), \\
\text{s.t.} ~~~~&h_{ij}(w_i,w_j)\leq 0, \quad(i, j) \in \mathcal{E},
\vspace{-3pt}  
\end{align*}
where $f_i$ denotes the local objectives on node $i$ and $h_{ij}$ encodes the constraints between adjacent nodes $i$ and $j$.

Specifically, a fundamental formulation in distributed optimization employs linear inequality constraints to couple decision variables across network nodes. Such formulation can be efficiently solved using IEQ-PDMM method \citep{heusdens2024distributed} and this can be formally expressed as
\begin{align}
\vspace{-3pt}  
\min_{\{w_i:i \in \mathcal{V}\}} ~~~~&\displaystyle \sum_{i \in \mathcal{V}} f_i(w_i), \label{eq.ineq}\\
\nonumber\text{s.t.} ~~~~&A_{i\mid j}w_i+A_{j\mid i}w_j\leq b_{i,j}, \quad(i, j) \in \mathcal{E}.
\vspace{-3pt}  
\end{align}
Constraints between entries are defined by $A_{i\mid j}$, $A_{j\mid i}$ and $b_{i,j}$.

\section{LAGO: Language Similarity-Aware Graph Optimization Framework}
Building upon the ALGEN paradigm and grounded in distributed optimization, we propose \textbf{LAGO} - a general framework for few-shot cross-lingual embedding inversion. LAGO operates in two stages: (1) constructing a language similarity graph to capture topological relationships between languages, and (2) solving a graph-constrained optimization problem to jointly estimate transformation matrices across languages. This section details both components and introduces two algorithmic variants that implement our optimization framework.

\begin{figure}[t!]
    \centering
\includegraphics[width=0.49\textwidth]{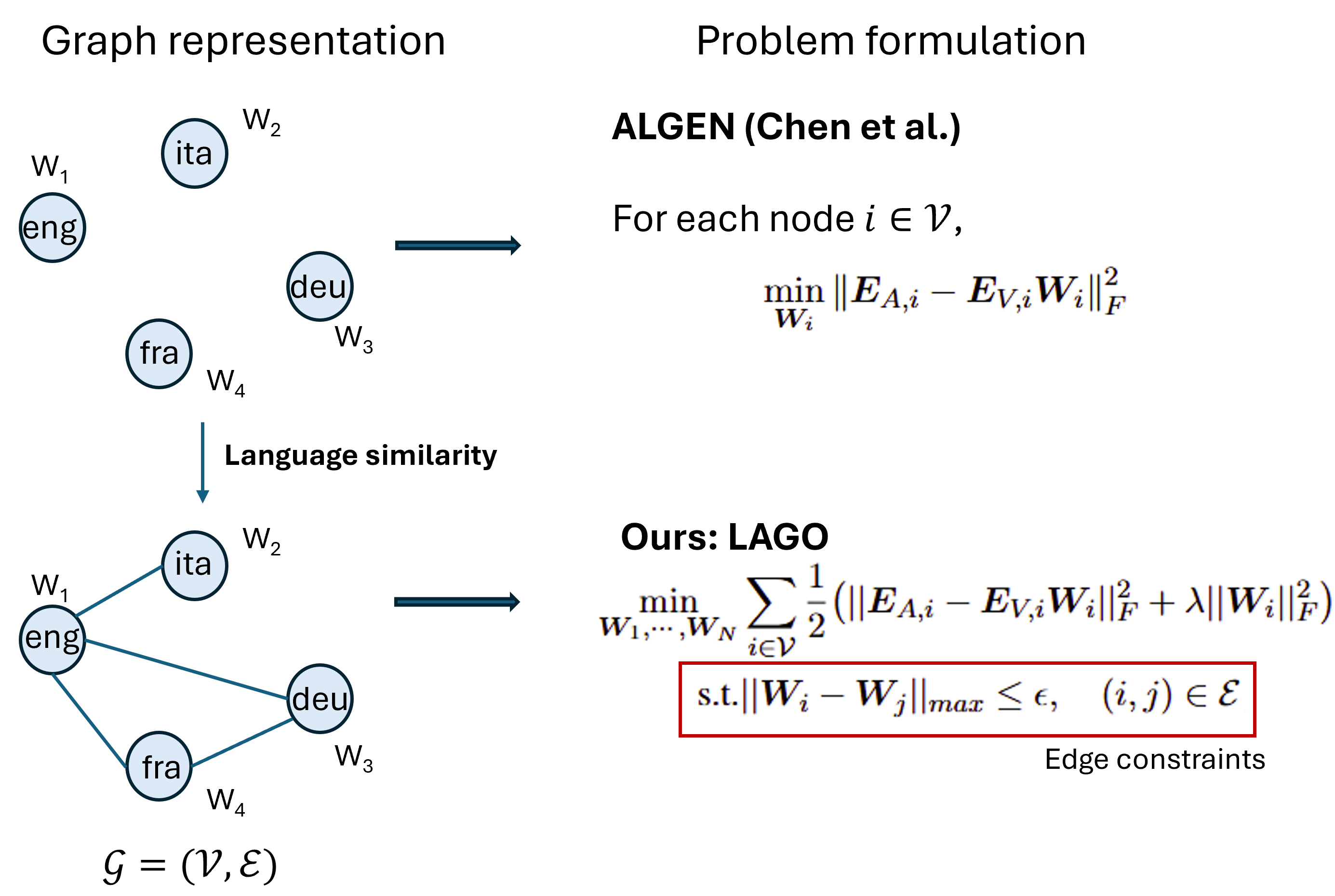}
    \caption{Illustration of LAGO vs. ALGEN~\citep{chen2025algen}. \textbf{Top}: ALGEN treats each language independently. \textbf{Bottom}: LAGO leverages language similarity by introducing edge constraints in a joint distributed optimization framework. }
    \label{fig:opti}
\end{figure}

\subsection{Step I: Language Similarity-Aware Graph Construction}
To formalize cross-lingual relationships, we propose to construct a linguistic topological graph $\mathcal{G} = (\mathcal{V}, \mathcal{E})$, where the set of nodes $\mathcal{V}$ represents languages and the set of undirected edges $\mathcal{E}$ encodes pairwise similarity. Language similarity is quantified using established metrics such as AJSP \citep{wichmann2022asjp} and Lang2vec \citep{littell2017uriel}. 
For a predefined threshold $r$, an edge is established between two languages $i$ and $j$ if their distance metric $D_{ij}<r$. 
Mathematically, given a distance matrix $\v D\in\mathbb{R}^{N\times N}$ over $N$ languages, the adjacency matrix $\v A \in\{0,1\}^{N\times N}$ of the resulting topology is derived as:
\vspace{-3pt}  
\[
\v A=\frac{1-sign(\v D-r)}{2}.
\vspace{-3pt}  
\]
See Appendix~\ref{app.topo} for a concrete example of graph construction.

\subsection{Step II: Graph-Constrained Optimization Algorithms}
Using the constructed graph, we reformulate the optimization objective, leveraging cross-lingual relationships, 
thereby enhancing embedding inversion attacks through knowledge transfer from linguistically related languages. 
In few-shot settings where local data is scarce, this formulation improves transferability by leveraging cross-lingual regularities. 
We introduce two optimization strategies: one enforcing hard constraints and one applying soft penalties. Let $\v W_i$ denote the transformation matrix at node $i$ (language $i$). To ensure stability in underdetermined settings (e.g., $b < m$), we incorporate Frobenius norm regularization to mitigate rank deficiency and enhance convergence.

\paragraph{Variant 1:  Linear Inequality Constraints}
The first algorithm variant introduces topological constraints to enforce consistency between adjacent nodes' transformation matrices. 
Formally, we formulate the objective as minimizing the sum of reconstruction errors across all nodes while imposing $\epsilon$-bounded constraints on the pairwise differences between neighbors' mapping matrices:
\begin{align*}
\vspace{-3pt}  
\min_{\v W_1, \cdots, \v W_N} & \displaystyle\sum_{i \in \mathcal{V}} \frac{1}{2}\big(||\v E_{A,i}-\v E_{V,i}\v W_i||_F^2+\lambda||\v W_i||_F^2\big), \\
\text{s.t.} & ||\v W_i-\v W_j||_{max} \leq \epsilon, \quad(i, j) \in \mathcal{E},
\vspace{-3pt}  
\end{align*}
where $\|\cdot\|_{max}$ denotes the entry-wise $\ell_\infty$ norm.  

This formulation corresponds to the general inequality-constrained form in Eq.~\eqref{eq.ineq}, where $A_{i \mid j} = -A_{j \mid i} = [1~ -1]^T$ and $b_{i,j} = [\epsilon~\epsilon]^T$. As such, it is compatible with the IEQ-PDMM optimization framework~\citep{heusdens2024distributed}. The update equations used in this framework are given below\footnote{The comparison in Eq.~\eqref{eq.z_update} is applied element-wise.}.
    \begin{align}
    \vspace{-3pt} 
    \nonumber\v W_i^{(t)}=&[\v E_{V,i}^\top \v E_{V,i}+(2cd_i+\lambda)\v I]^{-1}\\
    \nonumber&(\v E_{V,i}^\top \v E_{A,i}-\sum_{j \in \mathcal{N}_i}A_{i\mid j}\v Z_{i\mid j}^{(t)})\\
    \nonumber\v Y_{i\mid j}^{(t)}=&\v Z_{i\mid j}^{(t)}+2cA_{i\mid j}\v W_i^{(t)}-cb_{i, j}\\
    \v Z_{i\mid j}^{(t+1)}=&\left \{ \begin{array}{ll}
         \v Y_{j\mid i}^{(t+1)}, &\v Y_{i\mid j}^{(t+1)}+\v Y_{j\mid i}^{(t+1)}>0,\\ 
    -\v Y_{i\mid j}^{(t+1)}, &\text{otherwise,} \label{eq.z_update}
    \end{array}\right. 
    \vspace{-3pt} 
\end{align}
where $d_i=|\mathcal{N}_i|$ is the degree of node $i$. 

\paragraph{Variant 2: Total Variation Regularization}
The second variant introduces soft penalties using total variation across edges, a technique originally proposed for Byzantine-robust decentralized learning systems \citep{peng2021byzantine}. The optimization objective is formulated as follows:
\begin{align}
\vspace{-3pt} 
\nonumber\min_{\v W_1, \cdots, \v W_N} & \displaystyle\sum_{i \in \mathcal{V}} \big(\frac{1}{2}||\v E_{A,i}-\v E_{V,i}\v W_i||_F^2+\frac{\lambda}{2}||\v W_i||_F^2 \\
\nonumber&+\eta\sum_{j \in \mathcal{N}_i}||\v W_i-\v W_j||_{sum}\big),
\vspace{-3pt} 
\end{align}
where $\|\cdot\|_{sum}$ denotes the entry-wise $\ell_1$ norm. At time $t$, each node updates its $\v W$ with
\resizebox{0.98\linewidth}{!}{%
$\begin{aligned}
    \mathbf{W}_i^{(t+1)} &= \mathbf{W}_i^{(t)} - \frac{\alpha}{\sqrt{t+1}} \Big[ -\mathbf{E}_{V,i}^\top (\mathbf{E}_{A,i} - \mathbf{E}_{V,i} \mathbf{W}_i^{(t)}) \\
    &\quad + \lambda \mathbf{W}_i^{(t)} + \eta \sum_{j \in \mathcal{N}_i} \text{sign}(\mathbf{W}_i^{(t)} - \mathbf{W}_j^{(t)}) \Big],
\end{aligned}$
}
where $\alpha$ is the learning rate.

\subsection{Generalization: ALGEN as a Special Case}\label{ssec:equ}
Our proposed LAGO is general and subsumes prior method ALGEN as a special case. Specifically, in the inequality-constrained variant, when $\epsilon \to \infty$, cross-node constraints vanish, and each language node solves an independent alignment problem. Similarly, in the total variation setting, setting $\eta = 0$ decouples all nodes. In both cases, the optimization reduces to ALGEN’s per-language formulation with no cross-lingual structure. This highlights the flexibility of LAGO: by adjusting constraint strength, it interpolates between isolated optimization (as in ALGEN) and fully collaborative cross-lingual inversion. Our approach thus provides a principled, generalizable framework for multilingual attack design.

\begin{figure}[t!]
    \centering
\includegraphics[width=0.4\textwidth]{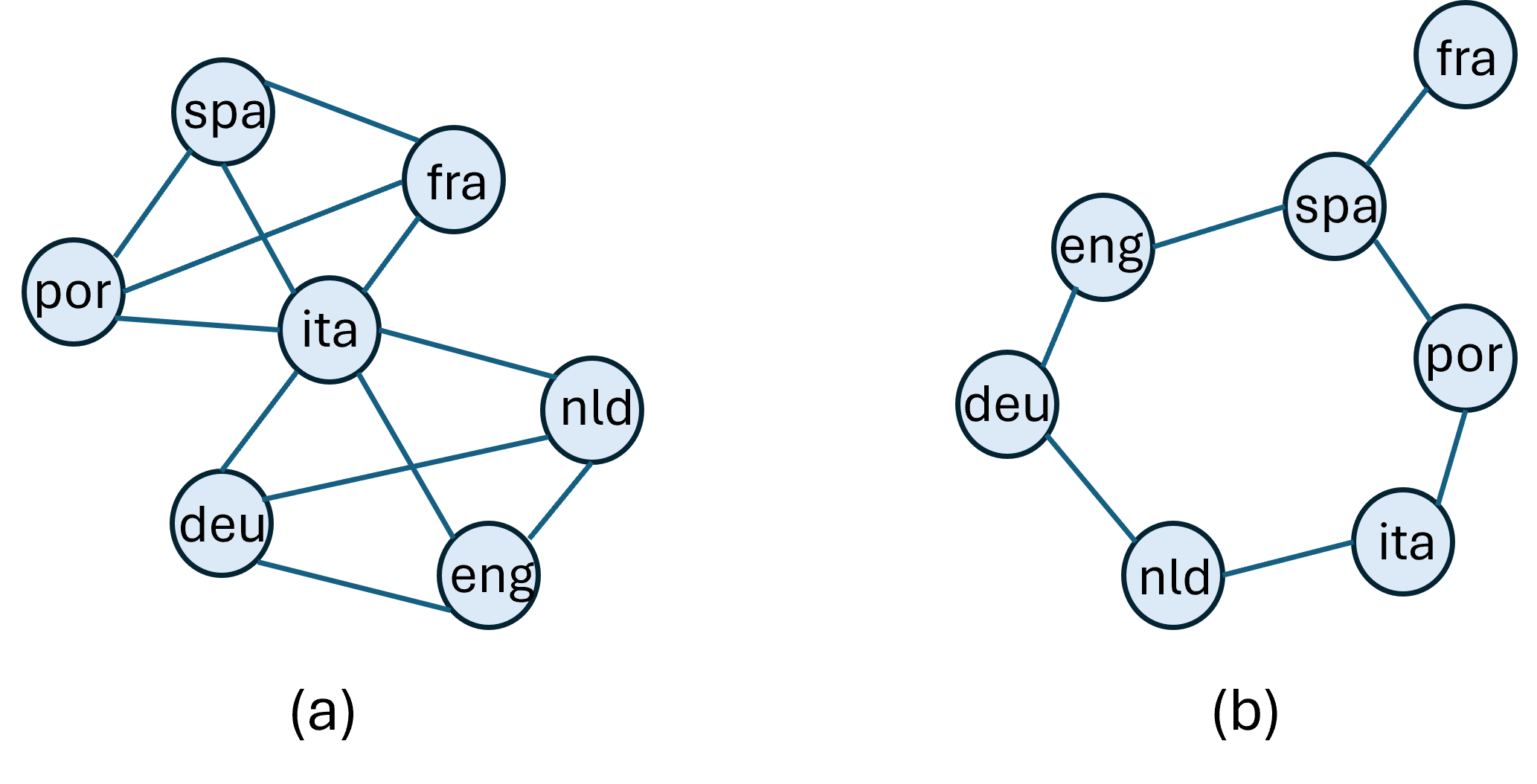}
    \caption{Example graphs using two Language Similarities: (a) AJSP model with $r=0.9$; (b) Lang2vec model with $r=0.45$.}
    \label{fig:ltg}
\end{figure}

\section{Experimental Setup}
\paragraph{Models and Dataset} Our attack framework is initialized using a pre-trained \textsc{Flan-T5} model. To evaluate the robustness of our approach, we conduct experiments with two distinct victim model encoders, \textsc{mT5}, \textsc{E5-small-v2} (\textsc{E5}) and OpenAI’s \textsc{text-embedding-ada-002} (\textsc{ada-2}) (see the details in Tabel~\ref{tab:llms}).
The decoder $dec_{A}(\cdot)$, fine-tuned on the \textsc{mMarco} English dataset \citep{bonifacio2021mmarco},
serves in this paper as \textit{the attack model} for simulating few-shot inversion attack scenarios.
We employ the current state-of-the-art ALGEN method as the baseline for the few-shot scenario, maintaining identical training and testing configurations for the decoder as those used in ALGEN.
To assess cross-lingual transferability, we select a subset of seven syntactically and lexically related languages: English, German, French, Dutch, Spanish, Italian and Portuguese. 

\paragraph{Language Graphs} We evaluate two distinct topologies derived from language similarities: AJSP and Lang2vec.
The tested topologies are illustrated in Fig.~ \ref{fig:ltg}.

\paragraph{Regularization Parameters}To accomplish substantial convergence, the number of iterations is fixed at 500. For the linear inequality constraints method, the convergence parameter is set to $c=0.4$, while for the TV penalty term method, the learning rate is chosen as $\alpha=0.01$. The computational cost of the attack is relatively low, using the topology of $7$ languages as an example, it takes approximately five minutes to compute a set of matrix $\{W_i:i\in\mathcal{V}\}$ with the inequality constrained formulation, while the total variation method is faster, completing the attack in about two minutes.

\paragraph{Evaluation Metrics}\label{app.eva}
We use Cosine similarity to measure the semantic alignment between the adversarial embeddings of the victim model $\v E_{V} \v W$ and the target attack embeddings $\v E_{A}$. Meanwhile, Rouge-L ~\citep{lin2004rouge} evaluates the lexical overlap between the reconstructed text and the ground truth by computing the length of their longest common subsequence, serving as a proxy for assessing the fidelity of the generated output at the lexical level.

\section{Analysis and Results}
To validate the effectiveness of our proposed \textbf{LAGO} framework, we experiment across a range of settings and tasks.
Each subsection addresses one research question, probing key aspects of cross-lingual transferability, generalization, and robustness to defense mechanisms \footnote{We open-source our code \url{https://anonymous.4open.science/r/ALGO_anonymous}.}.

\begin{figure}[t!]
    \centering
\includegraphics[width=0.45\textwidth]{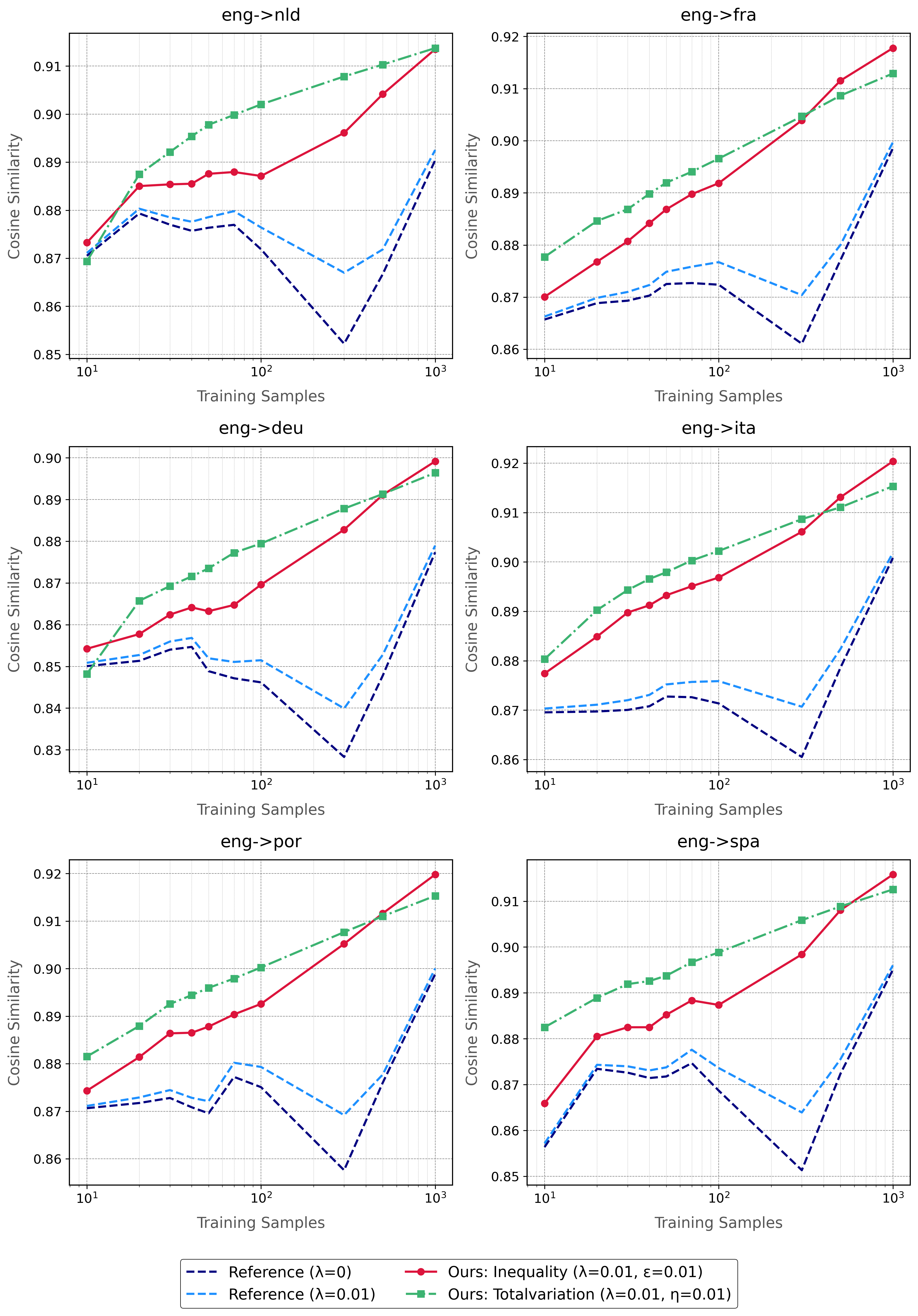}
    \caption{Cross-lingual Inversion Performances with AJSP Graph in Cosine Similarities across Training Samples.}
    \label{fig:cos}
\end{figure}

\begin{table*}[t!]
    \centering
     \resizebox{0.7\linewidth}{!}{
    \begin{tabular}{l|l|cccc|cccc}
    \toprule 
\multicolumn{2}{l|}{\textbf{Method}} & \multicolumn{4}{c|}{\textbf{Cosine Similarities}} & \multicolumn{4}{c}{\textbf{Rouge-L}} \\
\midrule

\multicolumn{2}{l|}{\textbf{Train Samples}} & 10 & 100 & 300& 1000 & 10 & 100& 300 & 1000 \\ \hline
\multirow{2}{*}{\rotatebox{90}{\textbf{\tiny{ALGEN}}}}  & -  & 0.8657 & 0.8723&0.8610 & 0.8986&10.07 &10.47& 10.22 &12.07 \\
& Reg.($\lambda=0.01$) & 0.8663& 0.8767&  0.8703& 0.8997& 10.14& 10.59& 10.37& 11.91\\
\midrule
\multirow{2}{*}{\rotatebox{90}{\textbf{\small{Ours}}}} & Inequality & 0.8701& 0.8919&0.9039 &\underline{0.9178}& 10.14& 11.09&\textbf{12.31} &\textbf{12.49}\\
& Total Variation & \underline{0.8777}& \underline{0.8966}& \underline{0.9046}& 0.9129& \textbf{10.87}& \textbf{11.59}& 11.46&12.30 \\
\bottomrule
\end{tabular}}

    \caption{Cross-lingual Inversion Performances of French embeddings with Attack Model trained in English in Cosine Similarities and Rouge-L scores across Training Samples. The best Rouge-L scores are \textbf{bold}, and the maximum cosine similarities are \underline{underlined}.}
    \label{tab:french_inversion}
\end{table*}

\subsection{Do Similar Languages Transfer Vulnerabilities?}
\begin{figure}[t!]
    \centering
\includegraphics[width=0.45\textwidth]{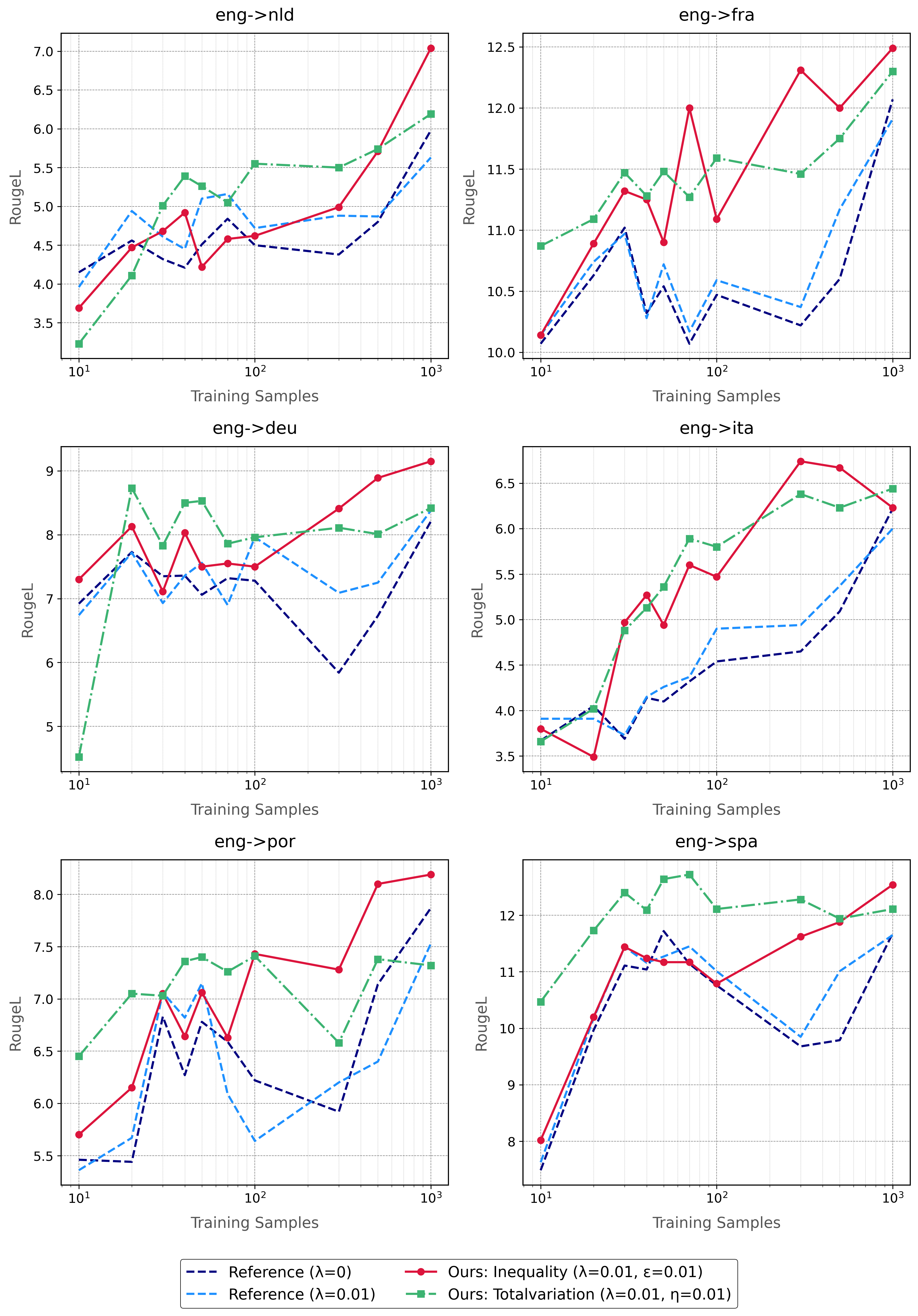}
    \caption{Cross-lingual Inversion Performances with AJSP Graph in Rouge-L Scores across Training Samples.}
    \label{fig:rougel}
\end{figure}

\begin{figure}[ht]
    \centering
\includegraphics[width=0.45\textwidth]{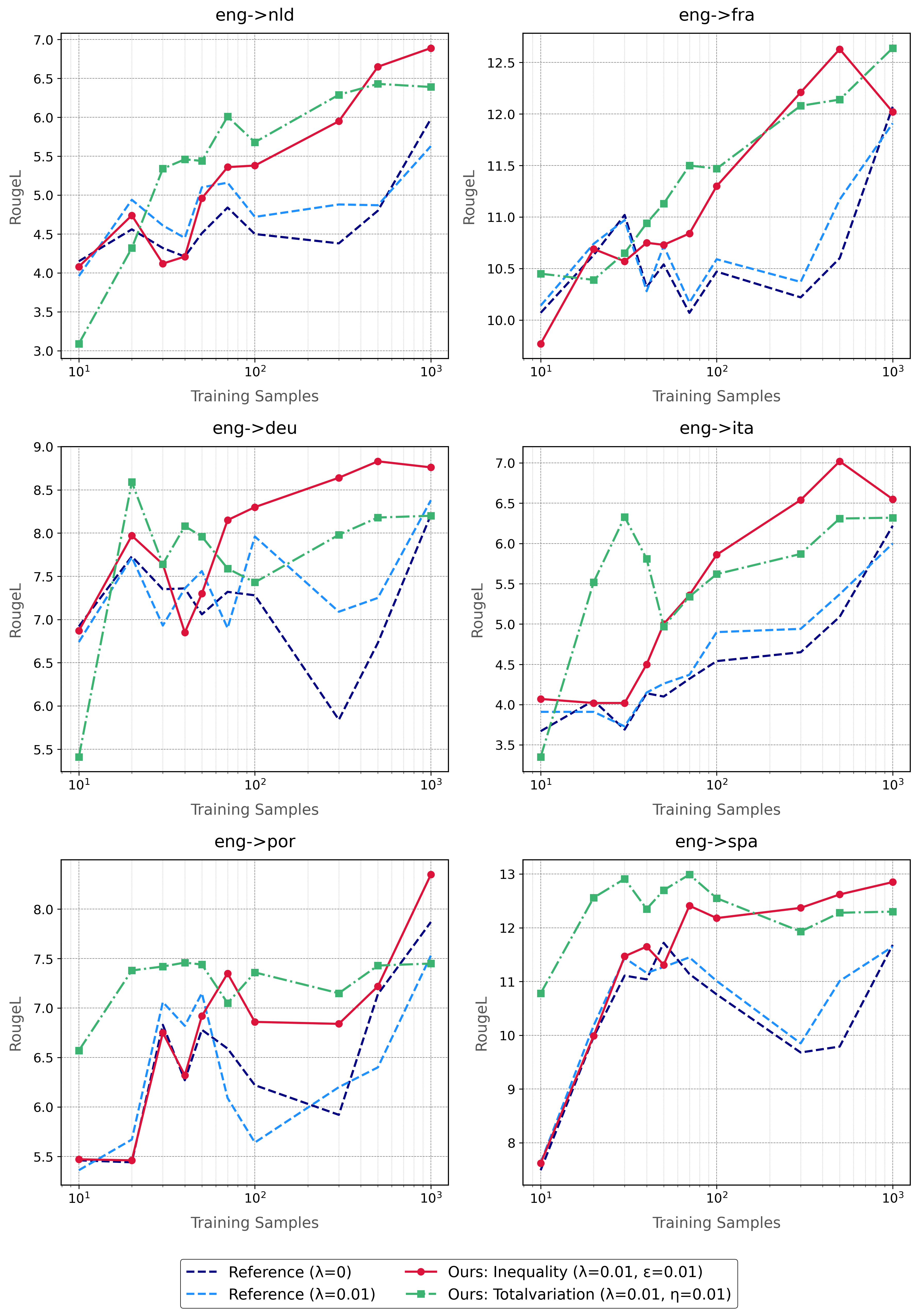}
    \caption{Cross-lingual Inversion Performances with Lang2vec Graph in Rouge-L Scores across Training Samples.}
    \label{fig:rougel_lang2vec}
\end{figure}

To assess whether language similarity aids attack transfer, we use an attack model trained on English data attack embeddings in other languages. 
We compare LAGO (with both optimization variants) to ALGEN baselines with and without Frobenius norm regularization ($\lambda = 0.01$), using 10 to 1000 training samples. Notice that the training sample is used exclusively for alignment.
For LAGO, we set $\epsilon = 0.01$ and $\eta = 0.01$.
As shown in Table~\ref{tab:french_inversion}, LAGO consistently improves both cosine similarity and Rouge-L scores on inverting French embeddings across all training sizes. In low-resource settings (e.g., 10 samples), our method yields a 10–20\% boost in Rouge-L over ALGEN. This trend generalizes to other languages, such as Dutch, German, Italian, Portuguese and Spanish, as demonstrated in Fig.~\ref{fig:cos};~\ref{fig:rougel} and Fig.~\ref{fig:rougel_lang2vec}.
These findings suggest that leveraging language similarity both mitigates data scarcity and optimizes cross-lingual generalization in low-resource settings.

\begin{figure}[ht]
    \centering
\includegraphics[width=0.45\textwidth]{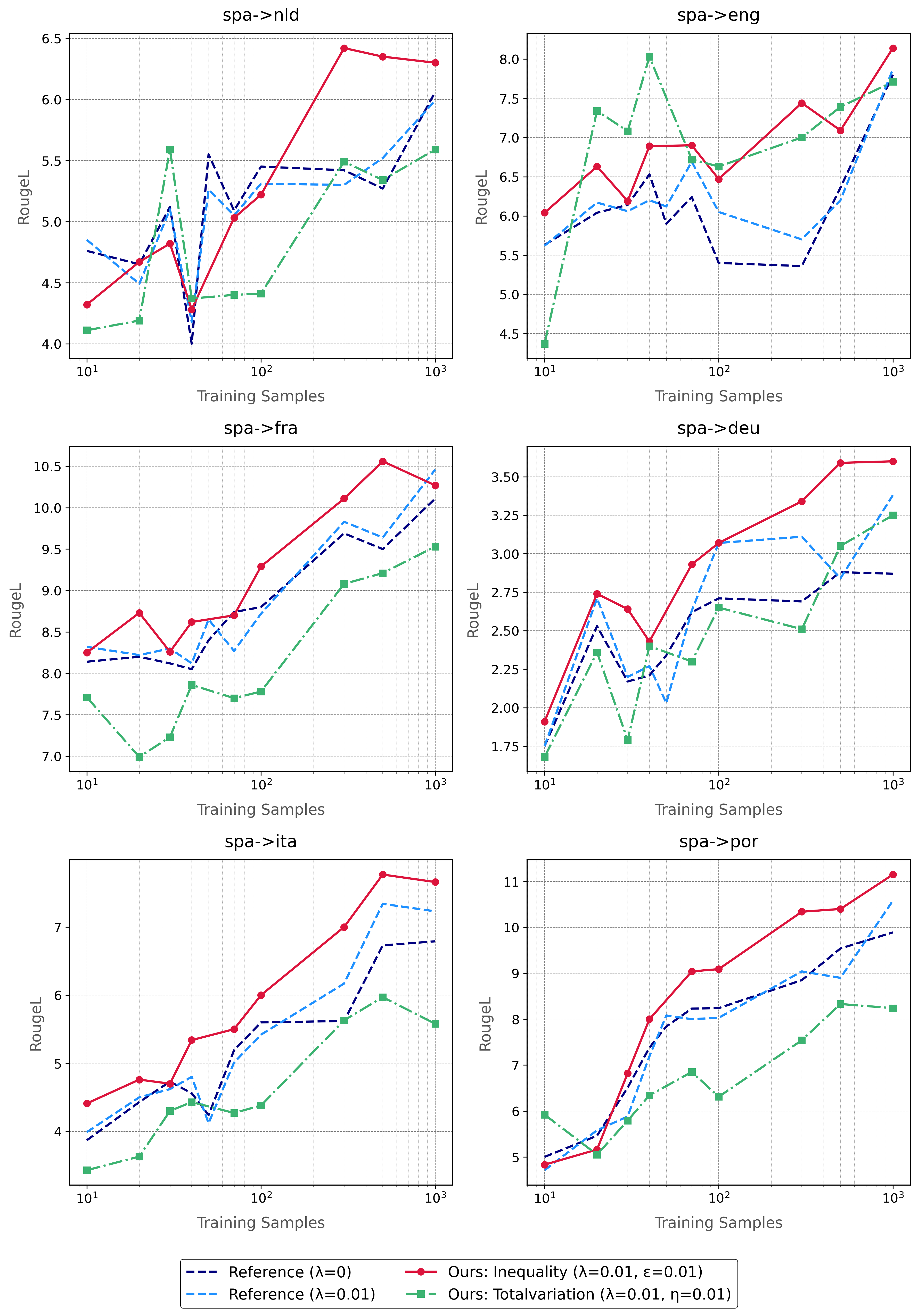}
    \caption{Cross-lingual Inversion Performances with Attack Model trained in Spanish in Rouge-L Scores across Training Samples.}
    \label{fig:rougel_spanish}
\end{figure}

\subsection{Does the choice of Language Similarity Metric Impact the Attack Effectiveness?}
To test the sensitivity of LAGO to the choice of language similarity measures,  we compare performance under two topologies: ASJP (lexical similarity) and Lang2vec (syntactic similarity). 
The results,  demonstrated in Fig. \ref{fig:rougel_lang2vec} and \ref{fig:cos_lang2vec} in Appendix \ref{other_result}, confirm that LAGO is robust to the choice of similarity metric. 
Lang2vec shows slightly better performance in moderate-data settings in terms of Rouge-L scores for moderately larger training sample sizes (>300). For instance, Dutch, with training samples of $|D_V|=500$, exhibits an increase from $5.71$ to $6.65$. Overall, our approach consistently outperforms the baseline in terms of attack efficacy, irrespective of the similarity metric. This suggests that LAGO is not contingent upon a specific language similarity framework but instead exhibits robust generalizability across diverse language structures.
Furthermore, the observed improvements in attack effectiveness indicate that our methodology is particularly advantageous for languages with shared linguistic features. Whether the similarity is lexical or syntactic, the attack remains effective, reinforcing its versatility.

\subsection{Is the Inversion Generalizable to Different Victim Models?}
We assess generalizability by evaluating our method on embeddings from \textsc{ada-2} and \textsc{E5} encoders. As shown in Appendix Fig.~\ref{fig:rougel_ada2} - \ref{fig:cos_e5}, LAGO consistently outperforms ALGEN in both cosine similarity and Rouge-L across these models.

Relatively, under the Rouge-L metric, the inequality constraint demonstrates stronger performance with larger sample sizes, whereas total variation proves more effective in extremely few-shot scenarios with fewer than 300 training samples. We attribute this to the flexibility of inequality constraints, a smaller sample size provides $\v W$ with greater degrees of freedom, thereby imposing relatively weaker restrictions on $\v W$ under the same $\epsilon$. Consequently, the performance of inequality constraints under smaller sample sizes aligns more closely with the ALGEN method. 

\subsection{Can other Languages assume the Source of Transfer?}
English as the most represented language in the pretrained LLMs, 
serves as an obvious choice for training the attack model to facilitate the inversion of other languages. 
We demonstrate that the proposed scheme remains robust even when the attack model is trained in an alternative language. As shown in Fig.~\ref{fig:rougel_spanish} and Fig.~\ref{fig:cos_spanish} in Appendix~\ref{other_result}, when Spanish is used as the attack language, LAGO continues to yield consistent improvements over the baseline. The cosine similarity increases across target languages, and the inequality-constrained variant shows stronger gains in Rouge-L, particularly under low-resource conditions.

We also observe performance disparities across specific language pairs. For example, the inversion performance from English to German is notably higher than that from Spanish to German - a pattern already present in the ALGEN baseline. This disparity may be attributed to two factors: differences in decoder training quality and variations in language similarity. In our constructed graph, English and German are directly connected (one-hop neighbors), whereas Spanish and German are two hops apart. The increased topological distance may weaken the effectiveness of parameter transfer, as similarity constraints exert less influence.

These observations suggest that the relative position of languages in the similarity graph - and not just data size or encoder choice - can influence transfer strength. Understanding the dynamics of language topology in transfer-based attacks presents an important direction for future work.

\subsection{Defenses}
We further investigate the effectiveness of differential privacy (DP) in mitigating embedding inversions. We employ the SNLI dataset~\citep{bowman-etal-2015-large} to fine-tune the decoder and subsequently transfer the adversarial attack framework to German, French and Spanish using the XNLI dataset~\citep{conneau2018xnlievaluatingcrosslingualsentence}. 
While the SNLI dataset is widely utilized for downstream tasks like text classification, \citet{chen2025algen} has demonstrated that with a strong privacy guarantee $\epsilon_{dp}=1$, model accuracy drops to 40\%, which is a significant reduction from the 60\% accuracy achieved at $\epsilon_{dp}=12$ where DP defenses show limited impact on utility and inversion performance.

In our setup, we apply two DP mechanisms: the Purkayastha Mechanism (PurMech) and the Normalized Planar Laplace Mechanism (LapMech) proposed by \citet{du2023sanitizing} in sentence embeddings. The privacy budget parameter is evaluated across $\epsilon_{dp}\in [1,4,8,12]$. As shown in Table~\ref{tab:snli_ldp_inequality}, Table~\ref{tab:snli_ldp_tv}, and Fig.~\ref{fig:cos_ep12}, \ref{fig:rougle_ep12} in Appendix~\ref{other_result}, inversion attacks in cross-lingual settings are highly sensitive to DP perturbations. Specifically, Rouge-L scores are consistently suppressed to below 2 across tested configurations. These results are consistent with theoretical expectations: more challenging examples, such as those in cross-lingual or low-resource settings, tend to be more sensitive to DP noise~\citep{carlini2019prototypical,feldman2020does}. While DP mechanisms provide meaningful protection against inversion, they incur a non-trivial utility cost, underscoring the need for more efficient, structure-aware defenses in multilingual NLP applications.

\begin{table}[htb!]
    \centering
     \resizebox{0.9\linewidth}{!}{
    \begin{tabular}{l|c|cc|cc}
    \toprule
   \multirow{2}{*}{\textbf{Lang}} & $\epsilon_{dp}$ & \textbf{Rouge-L}$\downarrow$ & \textbf{COS}$\downarrow$ & \textbf{Rouge-L}$\downarrow$ & \textbf{COS}$\downarrow$ \\ 
\cmidrule{2-6}   
         &  & \multicolumn{2}{c|}{LapMech} & \multicolumn{2}{c}{PurMech} \\ 
        \midrule 
          & 1 & 14.11 & 0.0017 & 14.05& 0.0156  \\ 
        eng $\rightarrow$ eng & 4 & 13.58& 0.0087& 13.94 &0.0348 \\ 
          & 8 &13.38 &0.0249 &13.45 & 0.0185\\ 
          & 12 & 13.90 &0.0345 & 12.77 &-0.0076 \\ 
                \midrule
          & 1 & 1.66 & -0.0013 & 1.31&  0.0136 \\ 
        eng $\rightarrow$ fra & 4 & 1.70& -0.0043& 1.58 & 0.0140\\ 
          & 8 & 1.42&0.0364 & 1.24&0.0166 \\ 
          & 12 & 1.60 & 0.0411&1.44  &0.0113 \\ 
                \midrule
          & 1 & 0.52 & -0.0119 & 0.49& 0.0090 \\ 
       eng $\rightarrow$ deu & 4 & 0.32&0.0065 & 0.54 &0.0127 \\ 
          & 8 & 0.62& 0.0187& 0.53& 0.0418\\ 
          & 12 &  0.44 &0.0327 & 0.43 &0.0367 \\  
                \midrule
          & 1 & 1.47 & 0.0062 &1.55 & -0.0090 \\ 
        eng $\rightarrow$ spa & 4 & 1.43&-0.0006 & 1.32 &0.0208 \\ 
          & 8 &1.70 & 0.0384 &1.35 &0.0266 \\ 
          & 12 & 1.52 &0.0160 & 1.41 &0.0389 \\  
        \bottomrule
    \end{tabular}
    }
     \caption{Cross-lingual Inversion Performance with $|D_V|$=100 on Classification Tasks on SNLI dataset with Local DP (Inequality). From a defender's perspective, $\downarrow$ means lower is better.}
         \label{tab:snli_ldp_inequality}
\end{table}

\section{Conclusion}
We proposed two optimization-based paradigms for enhancing few-shot crosslingual embedding inversions. Both are grounded in distributed optimization and operate over a topological network of languages constructed via language similarity. This graph structure enables collaborative alignment of embedding decoders, facilitating effective knowledge transfer even with extremely limited supervision.  Our experimental results show that both variants - linear inequality constraints and total variation penalties - consistently outperform existing methods, including ALGEN. In particular, the total variation approach demonstrates superior robustness in extremely few-shot settings, validating the importance of smooth cross-lingual parameter sharing. These findings establish language similarity as a key enabler of transferable inversion attacks, and underscore the need for privacy-preserving defenses that account for structural relationships among languages in multilingual NLP systems.
\newpage

\section*{Limitations}
While our approach outperforms prior methods, few-shot crosslingual embedding inversion remains a challenging task with substantial room for improvement. 
One limiting factor appears to be the decoder itself: even in the monolingual (original language) setting, inversion accuracy remains moderate, achieving approximately a 25 Rouge-L score on the \textsc{mMarco} English dataset with $|D_V|$=1k, and further declines under cross-lingual transfer. 
This suggests that the current attack decoder may struggle to generalize across languages, particularly when signal supervision is limited.

Interestingly, we observe that cross-lingual settings exhibit higher sensitivity to DP defenses, though such defenses incur significant utility degradation. This sensitivity highlights both the vulnerability and fragility of multilingual embeddings.
Future work could focus on enhancing the decoder training, e.g., through multilingual pretraining, or incorporating language-specific priors - which we expect could improve inversion performance in both monolingual and crosslingual scenarios.

\section*{Computational Resources}
We fine-tune the decoder on a single NVIDIA A40 GPU, with training completing in just three hours. Notably, ALGO operates with minimal GPU resource demands, enabling a true few-shot setup.

\section*{Ethics Statement}
We comply with the ACL Ethics Policy. 
The inversion attacks implemented in this paper can be misused and potentially harmful to proprietary embeddings.
We discuss and experiment with potential mitigation and defense mechanisms, and we encourage further research in developing effective defenses in this attack space.

\section*{Acknowledgements}
WY is founded by the EU ChipsJU and the Innovation Fund Denmark through the project CLEVER (no. 101097560); YC and JB are funded by the Carlsberg Foundation, under the Semper Ardens: Accelerate programme (project nr. CF21-0454). We further acknowledge the support of the AAU AI Cloud for providing computing resources.

\bibliography{custom}

\appendix

\newpage


\section{Derivation of Normal Equation} ~\label{normal_equation}
The optimal alignment matrix $\v W$ is obtained by minimizing a cost function $J$ that quantifies the discrepancy between the attack embedding matrix $\v E_{A}$ and the transformed victim embeddings $\v E_{V\rightarrow A}=\v E_{V} \v W$: 

\begin{equation}
\begin{aligned}
J(\v W) &= \frac{1}{2} (\v E_A  - \v E_V \v W)^{T} (\v E_A - \v E_V \v W) \\
& = \frac{1}{2}(\v E_A^{T} \v E_A - \v E_A^{T} \v E_V \v W -  (\v E_{V} \v W)^{T} \v E_{A} \\
& + (\v E_{V} \v W)^{T} \v E_V \v W) \\
& = \frac{1}{2}(\v E_A^{T} \v E_A - \v E_A^{T} \v E_V \v W -   \v W^{T}\v E_{V}^{T} \v E_{A} \\
& +  \v W^{T}\v E_{V}^{T} \v E_V \v W).
\end{aligned}
\end{equation}
By calculating the derivatives of $J(\v W)$, we have
\begin{equation}
    \begin{aligned}
        \nabla_{\v W} J(\v W)  & =\frac{1}{2} \nabla_{\v W} (\v E_A^{T} \v E_A - \v E_A^{T} \v E_V \v W \\
        & - \v W^{T}\v E_{V}^{T} \v E_{A} +  \v W^{T}\v E_{V}^{T} \v E_V \v W) \\
        & = 2\v E_{V}^{T} \v E_V \v W -2 \v E_{V}^{T} \v E_{A} .
    \end{aligned}
\end{equation}
The optimized $\v W$ is achieved when the derivative is equal to 0,
\begin{equation}
    \v E^{T}_{V} \v E_{V} \v W = \v E^{T}_{V} \v E_{A}.
\end{equation}
Then, the matrix $\v W$ that minimizes $J(\v W)$ is
\begin{equation}
\v W = (\v E_V^{T}\v E_V)^{-1}\v E^{T}_{V} \v E_{A}.
\end{equation}

\section{Topology Construction}\label{app.topo}
\begin{figure}[ht]
    \centering
\includegraphics[width=0.35\textwidth]{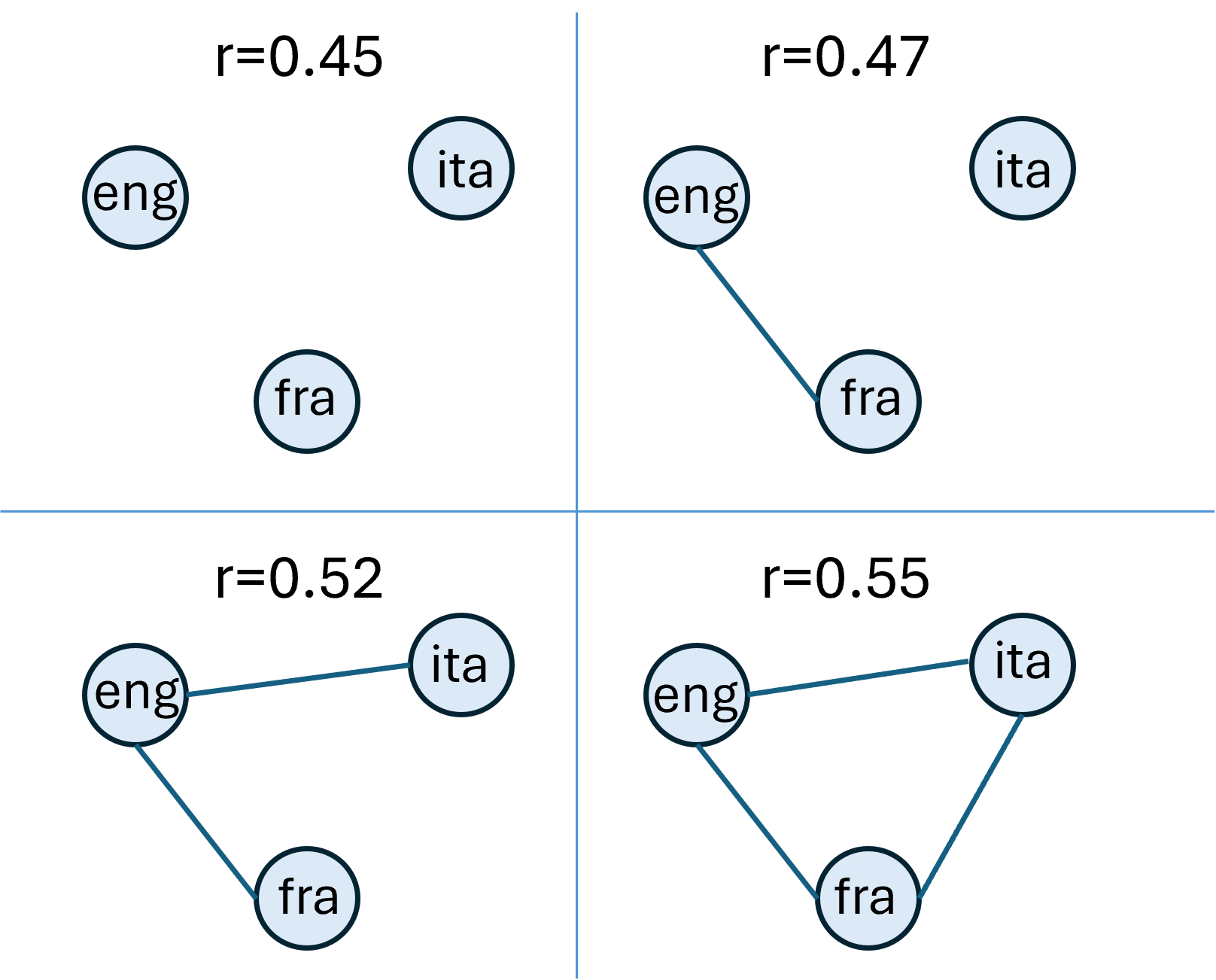}
    \caption{Linguistic topological graph of English, French and Italian with different threshold $r$. The higher the threshold, the denser the connectivity.}
    \label{fig:topo}
\end{figure}

To illustrate this approach, consider the syntactic distance matrix obtained from Lang2vec for English (eng), French (fra), and Italian (ita):
$$\v D=\begin{bmatrix}
0 & 0.46 & 0.51\\
0.46 & 0 & 0.55 \\
0.51 & 0.55 & 0
\end{bmatrix}$$
where each entry $D_{ij}$ represents the syntactic dissimilarity between language pairs. By applying different threshold values $r$, we construct distinct topological configurations of language relationships. Fig.~\ref{fig:topo} demonstrates how the network connectivity varies with increasing $r$ values, revealing:
\begin{itemize}
    \item At $r = 0.45$: No edges form
    \item At $r = 0.47$: eng-fra connection emerges
    \item At $r = 0.52$: eng-ita connection appears while fra-ita remains disconnected
    \item At $r = 0.56$: Complete graph forms
\end{itemize}

\section{Other Experimental Results} ~\label{other_result}

\begin{table*}[t!]
    \centering
    \resizebox{\linewidth}{!}{
    \begin{tabular}{l|l|l|l|l}
    \toprule
    Model  &  Huggingface & Architecture & \#Languages &   Dimension\\
    \midrule
       \textsc{Flan-T5}~\citep{chung2022scalinginstructionfinetunedlanguagemodels}  & google/flan-t5-small  & Encoder-Decoder  & 60  & 512 \\
       ~\textsc{E5-small-v2}~\citep{wang2022text}& intfloat/e5-small-v2 &  Encoder & 1 & 384 \\
      ~\textsc{mT5}~\citep{xue2020mt5} & google/mt5-base &   Encoder-Decoder & 102  &768\\
    ~\textsc{text-embedding-ada-002}& OpenAI API &  Encoder & 100+ & 1536 \\
    \bottomrule  
    \end{tabular}}
    \caption{Details of LLMs and Embeddings.}
    \label{tab:llms}
\end{table*}

\begin{figure}[ht]
    \centering
\includegraphics[width=0.45\textwidth]{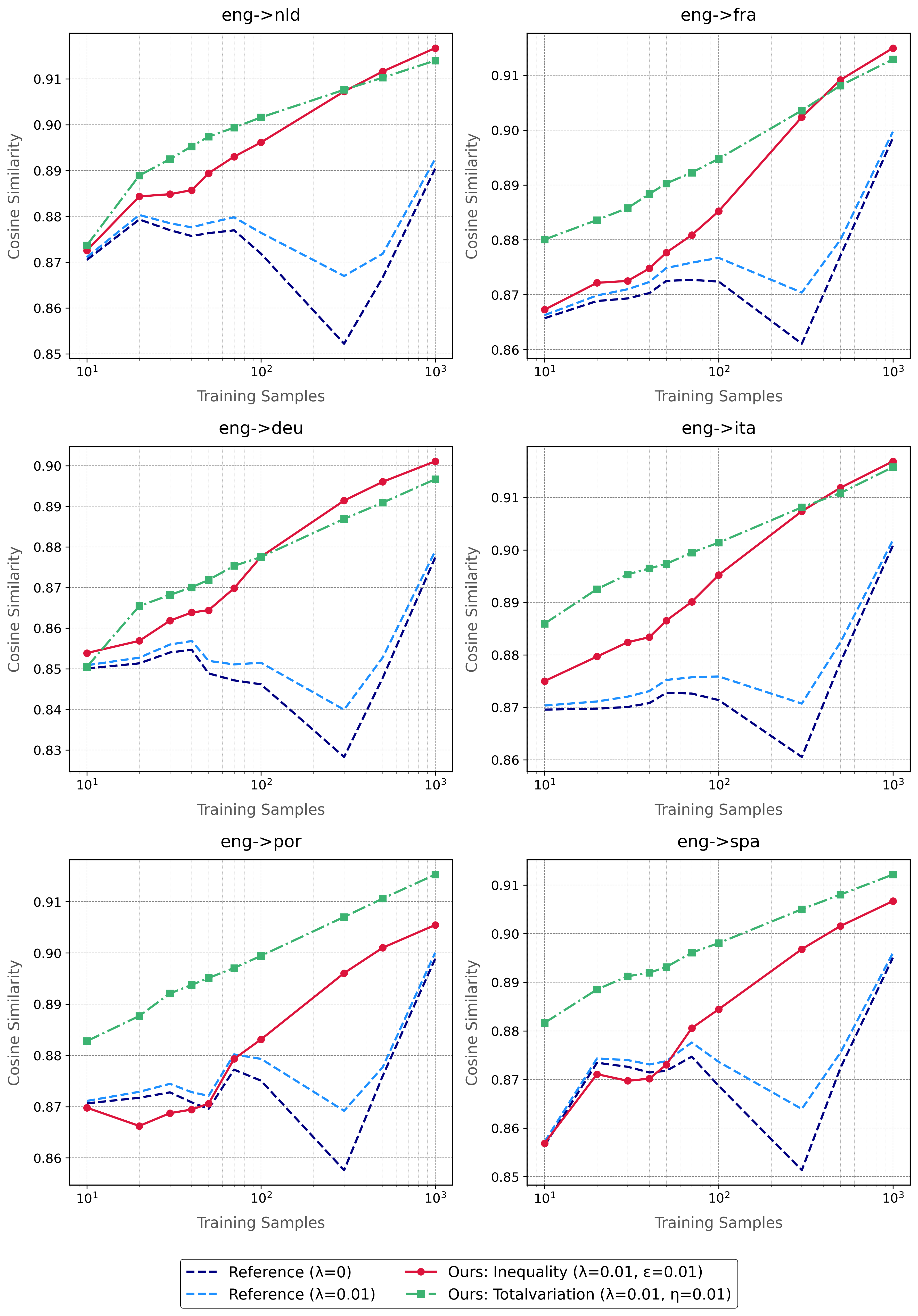}
    \caption{Cross-lingual Inversion Performances with Lang2vec Graph in Cosine Similarities across Training Samples.}
    \label{fig:cos_lang2vec}
\end{figure}

\begin{figure}[ht]
    \centering
\includegraphics[width=0.45\textwidth]{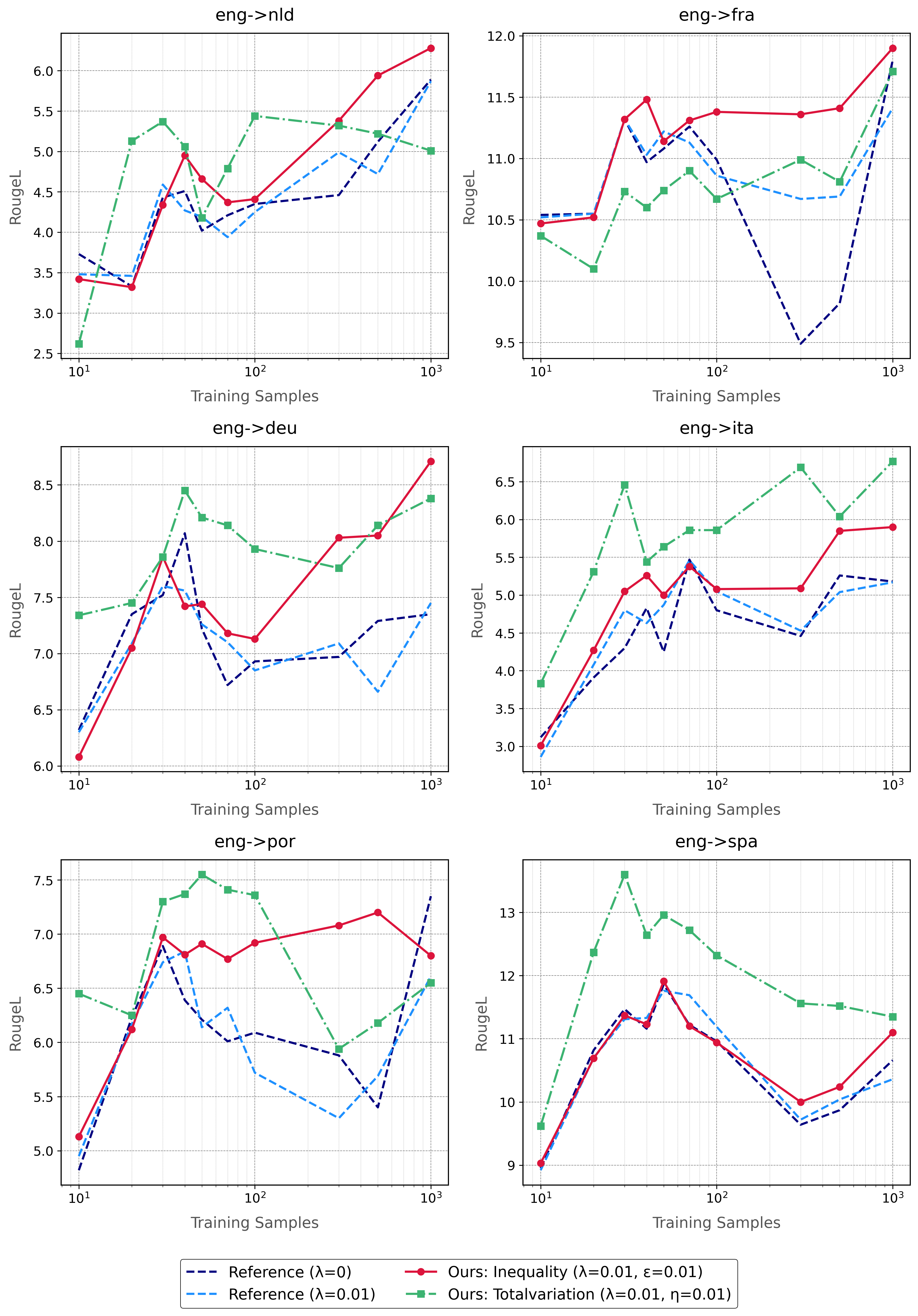}
    \caption{Cross-lingual Inversion Performances with ADA-2 Victim Model in Rouge-L Scores across Training Samples.}
    \label{fig:rougel_ada2}
\end{figure}

\begin{figure}[ht]
    \centering
\includegraphics[width=0.45\textwidth]{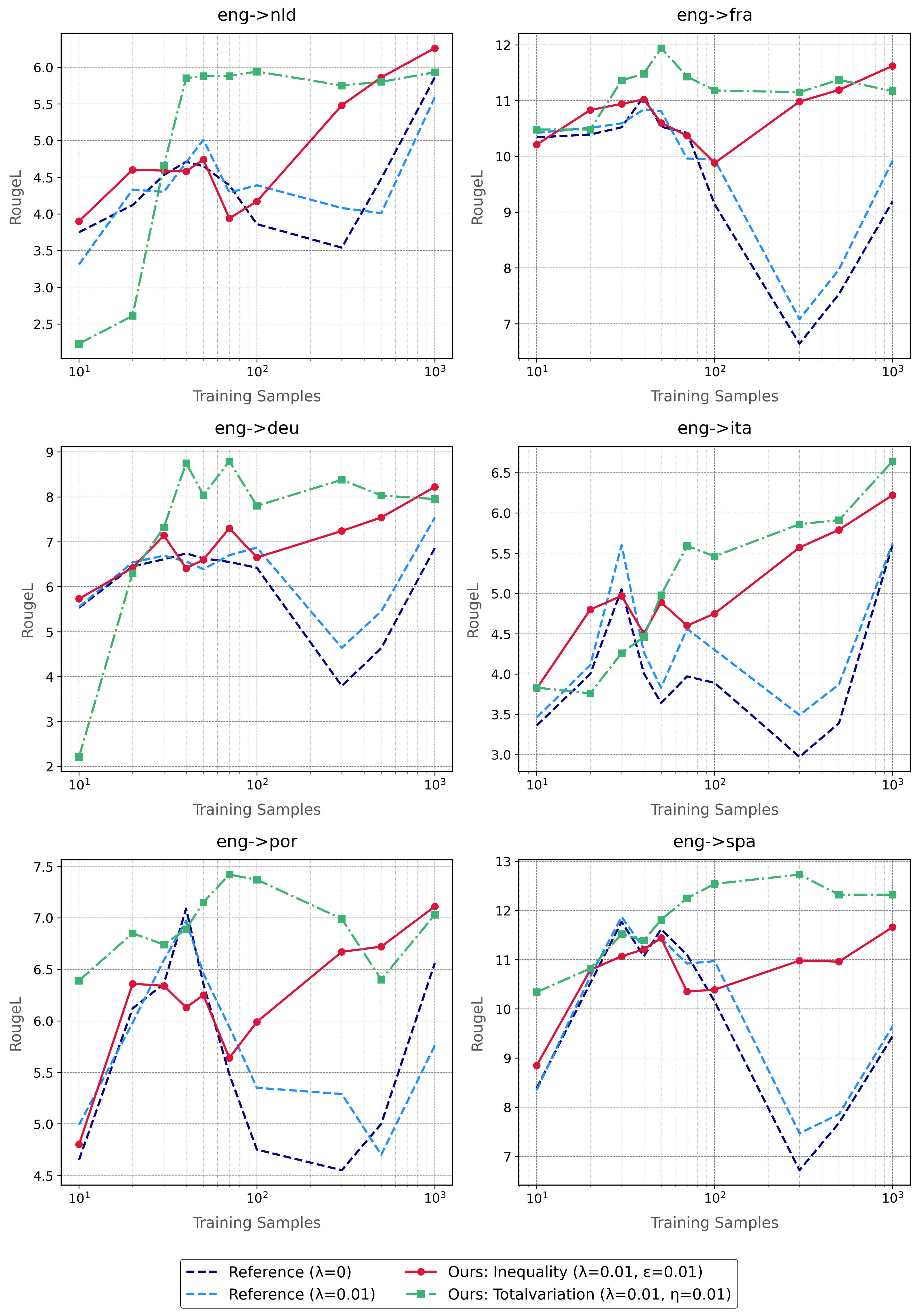}
    \caption{Cross-lingual Inversion Performances with E5 Victim Model in Rouge-L Scores across Training Samples.}
    \label{fig:rougel_e5}
\end{figure}

\begin{figure}[ht]
    \centering
\includegraphics[width=0.45\textwidth]{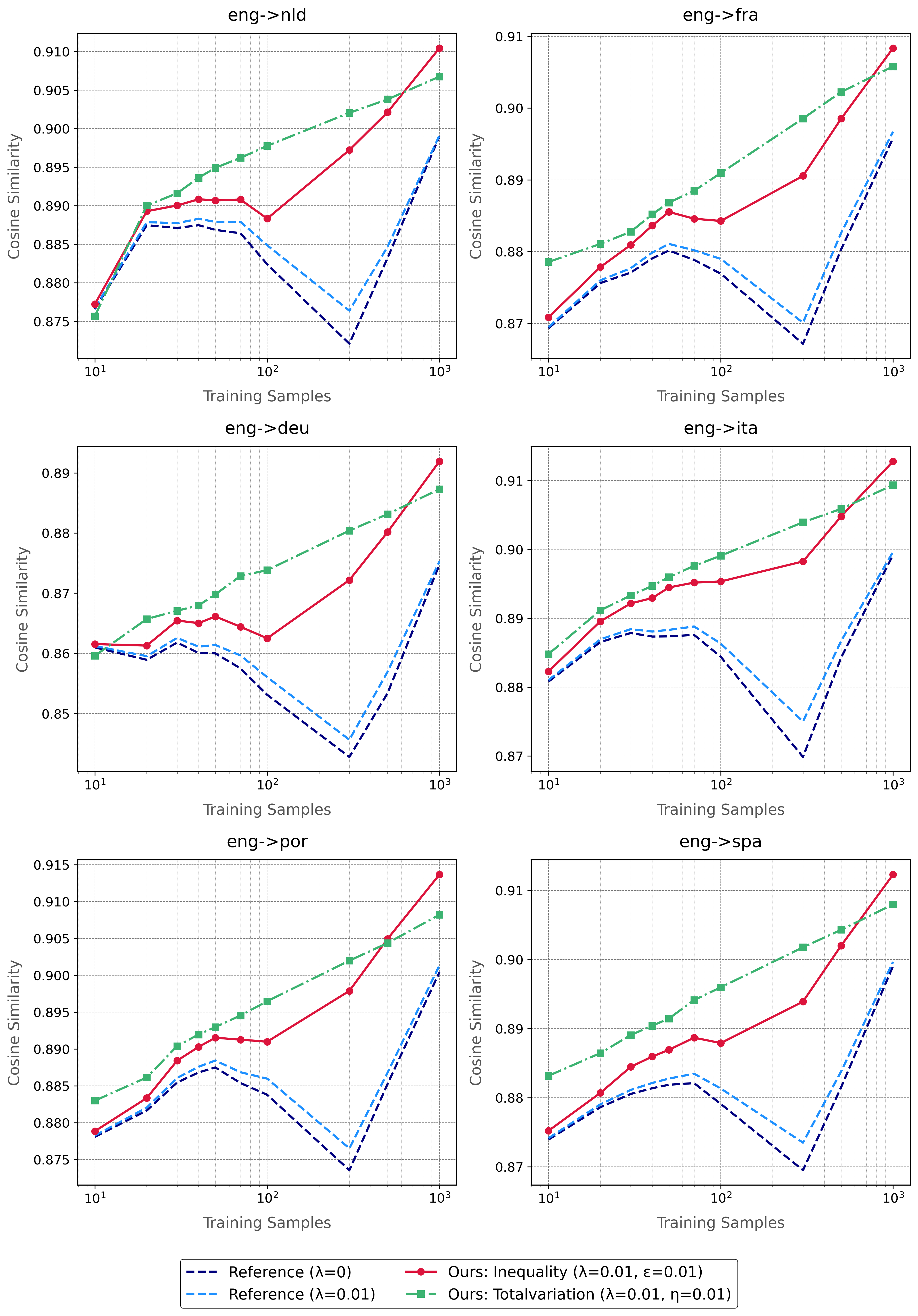}
    \caption{Cross-lingual Inversion Performances with ADA-2 Victim Model in Cosine Similarities across Training Samples.}
    \label{fig:cos_ada2}
\end{figure}

\begin{figure}[ht]
    \centering
\includegraphics[width=0.45\textwidth]{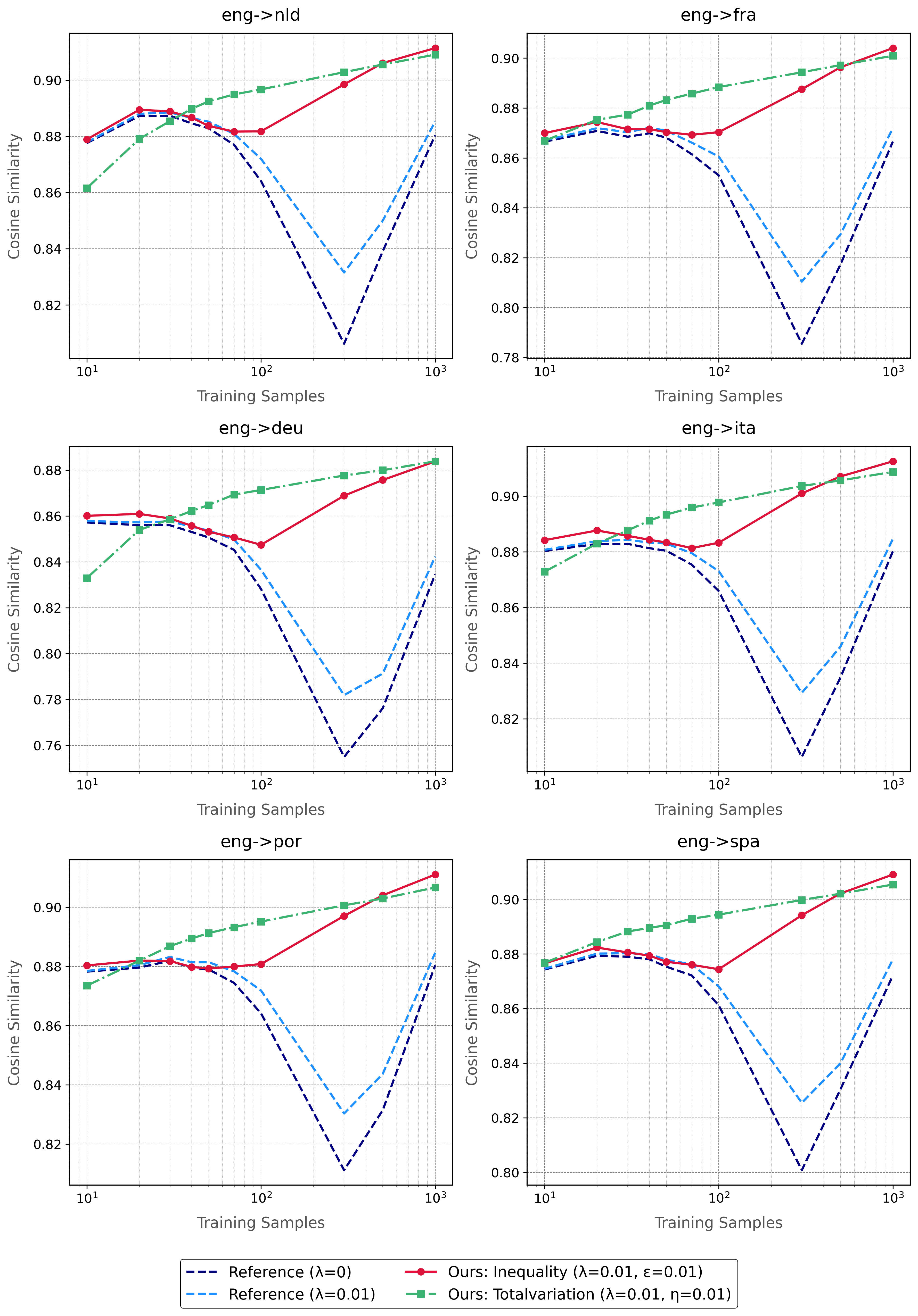}
    \caption{Cross-lingual Inversion Performances with E5 Victim Model in Cosine Similarities across Training Samples.}
    \label{fig:cos_e5}
\end{figure}

\begin{figure}[ht]
    \centering
\includegraphics[width=0.45\textwidth]{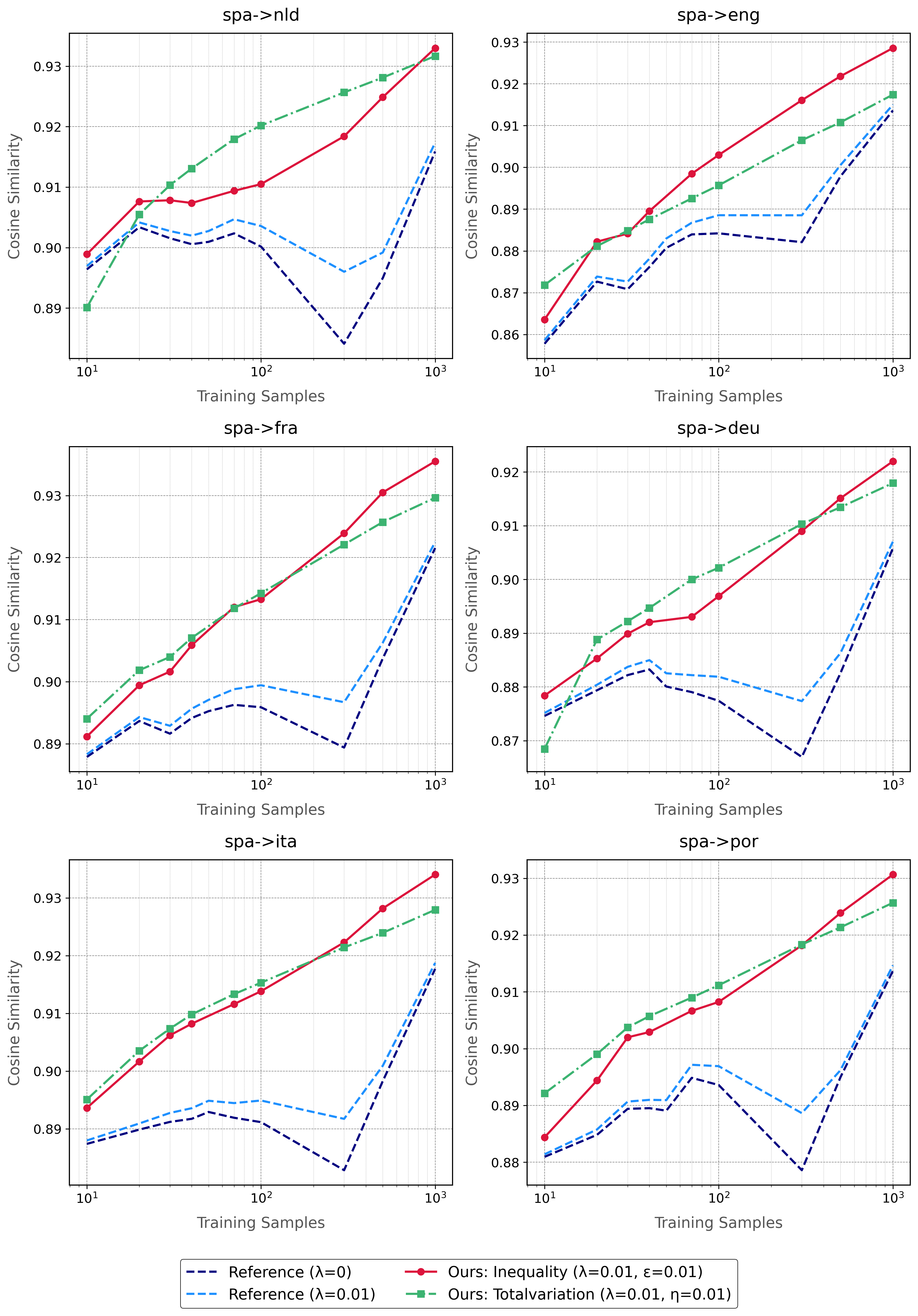}
    \caption{Cross-lingual Inversion Performances with Attack Model trained in Spanish in Cosine Similarities across Training Samples.}
    \label{fig:cos_spanish}
\end{figure}

\begin{table}[htb!]
    \centering
     \resizebox{0.9\linewidth}{!}{
    \begin{tabular}{l|c|cc|cc}
    \toprule
   \multirow{2}{*}{\textbf{Lang}} & $\epsilon_{dp}$ & \textbf{Rouge-L}$\downarrow$ & \textbf{COS}$\downarrow$ & \textbf{Rouge-L}$\downarrow$ & \textbf{COS}$\downarrow$ \\ 
\cmidrule{2-6}   
         &  & \multicolumn{2}{c|}{LapMech} & \multicolumn{2}{c}{PurMech} \\ 
        \midrule 
          & 1 & 13.16 & 0.0751& 13.35& 0.0199\\ 
        eng->eng & 4 & 12.95& 0.0257& 12.61 &0.0510 \\ 
          & 8 & 14.01&0.0845 & 13.88&0.1320 \\ 
          & 12 & 13.52 & 0.1720& 13.86 &0.1162 \\ 
                \midrule
          & 1 & 1.60 &-0.0168& 1.90 & -0.0189 \\ 
        eng->fra & 4 &1.77 &-0.0161 & 2.10 &0.1081 \\ 
          & 8 &2.02 &0.1040 & 2.10&0.1428 \\ 
          & 12 &1.92  &0.1271 & 2.46 & 0.1853\\ 
                \midrule
          & 1 & 0.86 & 0.0080 &0.62 & -0.0240 \\ 
       eng->deu & 4 & 0.99&0.0259 & 0.62 & -0.0216\\ 
          & 8 &0.77 &0.0960 & 0.64&0.0881 \\ 
          & 12 & 0.70 &0.1815 & 1.22 & 0.1944\\  
                \midrule
          & 1 & 1.58 &0.0431 & 1.78 &0.0729  \\ 
        eng->sap & 4 & 1.35& 0.0318& 1.45 &0.0360 \\ 
          & 8 &1.87 &0.2408 & 1.94&0.1119 \\ 
          & 12 & 1.65 & 0.1875& 2.29 & 0.1846\\  
        \bottomrule
    \end{tabular}
    }
     \caption{Cross-lingual Inversion Performance with $|D_V|$=100 on Classification Tasks on SNLI dataset with Local DP (Total Variation). From a defender's perspective, $\downarrow$ means lower are better.}
         \label{tab:snli_ldp_tv}
\end{table}

\begin{figure}[ht]
    \centering
\includegraphics[width=0.45\textwidth]{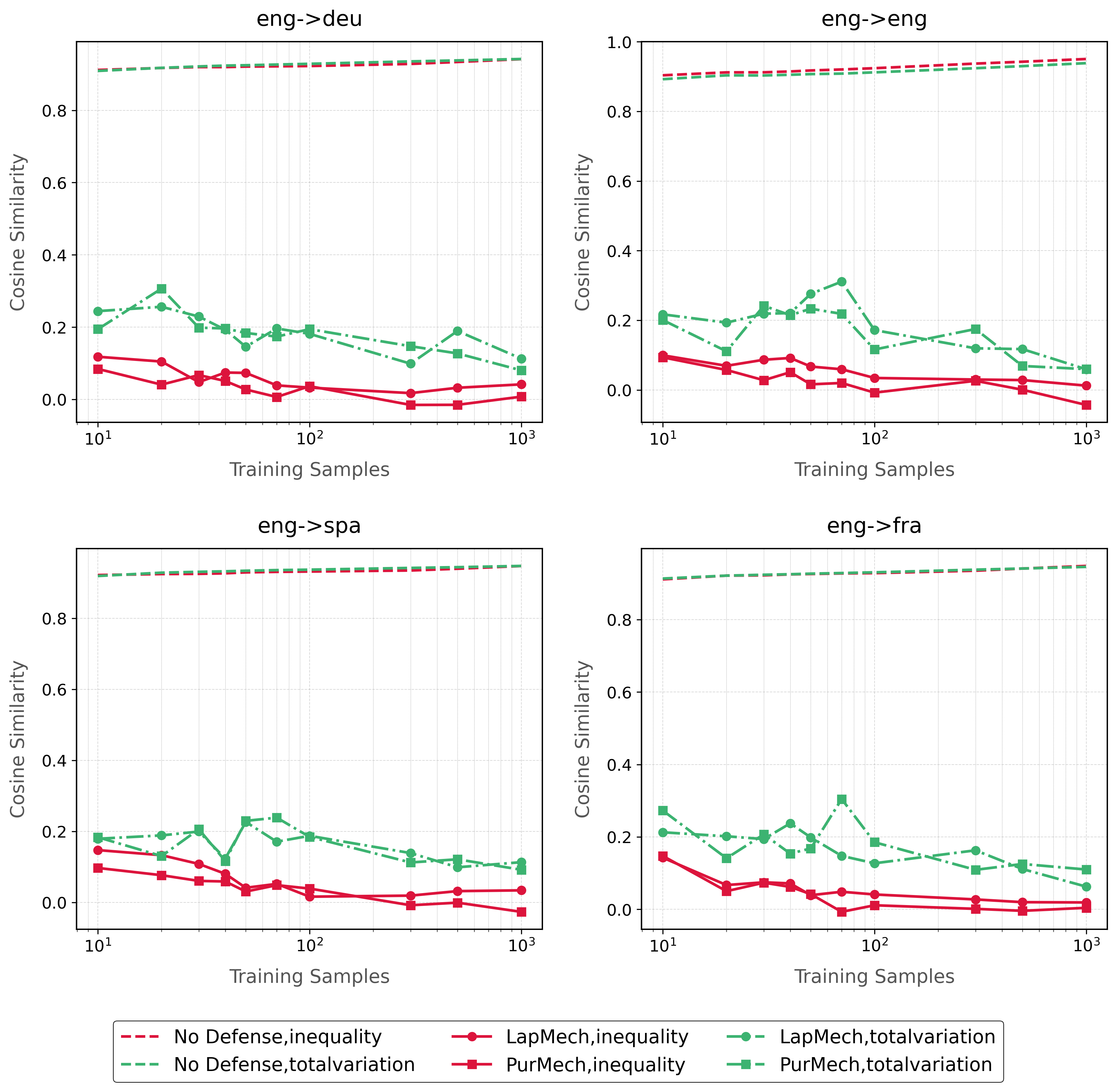}
    \caption{Cross-lingual Inversion Performance on Classification
Tasks on SNLI dataset with Local DP ($\epsilon_{dp}=12$) in Cosine Similarities.}
    \label{fig:cos_ep12}
\end{figure}

\begin{figure}[ht]
    \centering
\includegraphics[width=0.45\textwidth]{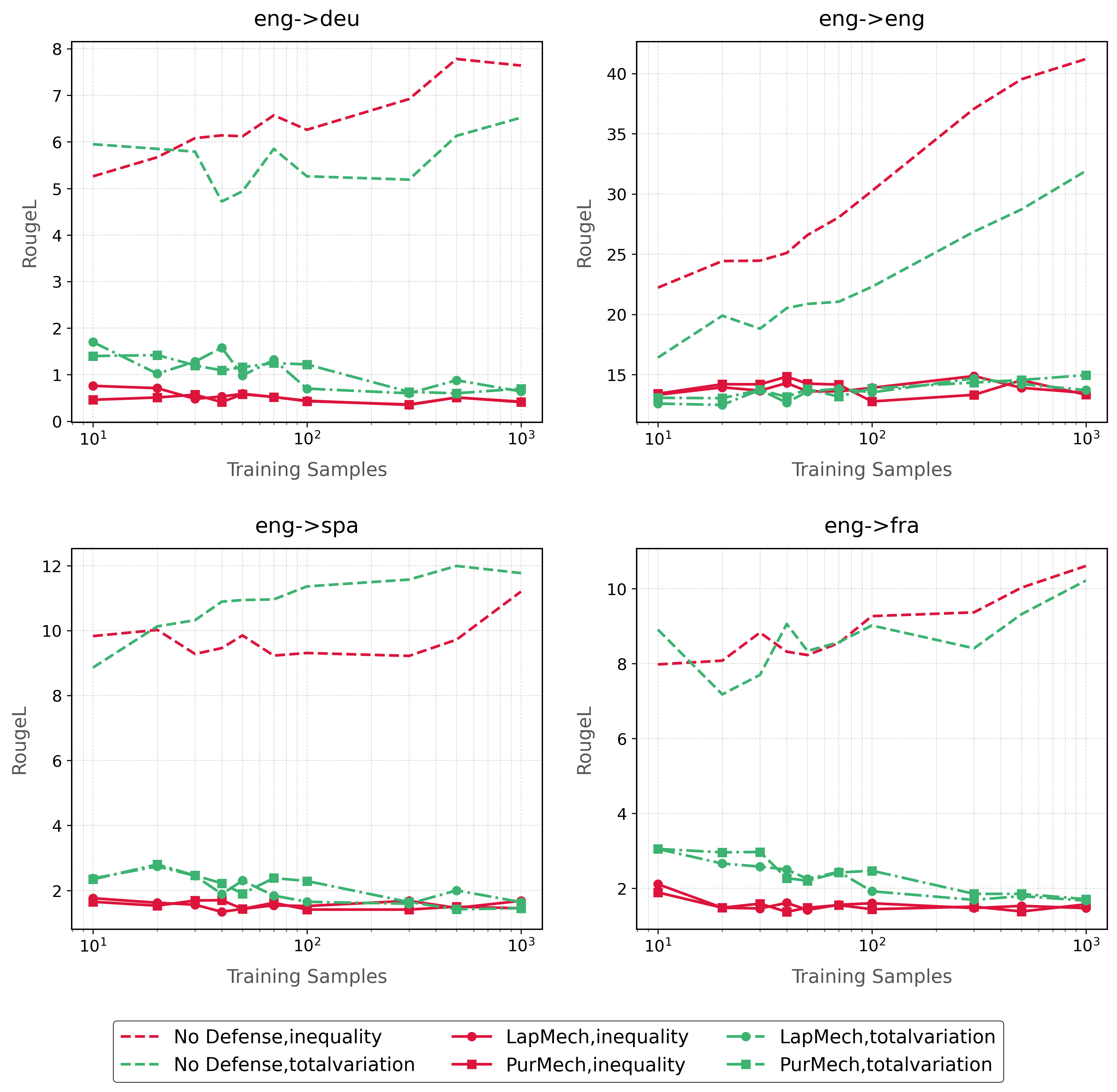}
    \caption{Cross-lingual Inversion Performance on Classification
Tasks on SNLI dataset with Local DP ($\epsilon_{dp}=12$) in Rouge-L Scores.}
    \label{fig:rougle_ep12}
\end{figure}

\end{document}